\documentclass[conference]{IEEEtran}
\IEEEoverridecommandlockouts
\usepackage{cite}
\usepackage{url}
\usepackage[utf8]{inputenc}
\usepackage{booktabs,multirow}
\usepackage{amsmath,amssymb,amsfonts}
\usepackage{algorithm}
\usepackage{algorithmicx,eqparbox,array}
\usepackage[noend]{algpseudocode}
\algnewcommand{\LineComment}[1]{\State // #1}
\usepackage{graphicx}
\usepackage{textcomp}
\usepackage{xcolor}
\def\BibTeX{{\rm B\kern-.05em{\sc i\kern-.025em b}\kern-.08em
    T\kern-.1667em\lower.7ex\hbox{E}\kern-.125emX}}
    
\newcommand\blfootnote[1]{%
  \begingroup
  \renewcommand\thefootnote{}\footnote{#1}%
  \addtocounter{footnote}{-1}%
  \endgroup
}
    
\begin{document}

\title{Graph Colouring Meets Deep Learning: Effective Graph Neural Network Models for Combinatorial Problems
\thanks{This research was partly supported by Coordenação de Aperfeiçoamento de Pessoal de Nível Superior (CAPES) - Finance Code 001 and by the Brazilian Research Council CNPq.}
}

\author{\IEEEauthorblockN{Henrique Lemos\IEEEauthorrefmark{1},
Marcelo Prates\IEEEauthorrefmark{2}, Pedro Avelar\IEEEauthorrefmark{3} and
Luis Lamb\IEEEauthorrefmark{4}}
\IEEEauthorblockA{Institute of Informatics,
Federal University of Rio Grande do Sul\\
Porto Alegre, Brazil\\
Email: \IEEEauthorrefmark{1}hlsantos@inf.ufrgs.br,
\IEEEauthorrefmark{2}morprates@inf.ufrgs.br,
\IEEEauthorrefmark{3}phcavelar@inf.ufrgs.br,
\IEEEauthorrefmark{4}lamb@inf.ufrgs.br}}

\maketitle

\begin{abstract}
Deep learning has consistently defied state-of-the-art techniques in many fields over the last decade. However, we are just beginning to understand the capabilities of neural learning in symbolic domains. Deep learning architectures that employ parameter sharing over graphs can produce models which can be trained on complex properties of relational data. These include highly relevant $\mathcal{NP}$-Complete problems, such as SAT and TSP. In this work, we showcase how Graph Neural Networks (GNN) can be engineered -- with a very simple architecture -- to solve the fundamental combinatorial problem of graph colouring. Our results show that the model, which achieves high accuracy upon training on random instances, is able to  generalise to graph distributions different from those seen at training time. Further, it performs better than the Neurosat, Tabucol and greedy baselines for some distributions. In addition, we show how vertex embeddings can be clustered in multidimensional spaces to yield constructive solutions even though our model is only trained as a binary classifier. In summary, our results contribute to shorten the gap in our understanding of the algorithms learned by GNNs, as well as hoarding empirical evidence for their capability on hard combinatorial problems. Our results thus contribute to the standing challenge of integrating robust learning and symbolic reasoning in Deep Learning systems.
\end{abstract}


\section{Introduction}
 Deep Learning (DL) models have defied several state-of-the-art techniques in tasks such as image recognition \cite{krizhevsky2012imagenet,simonyan2014very,li2015convolutional} and natural language processing \cite{cho2014properties,bahdanau2014neural}. Deep Learning has also been blended with Reinforcement Learning algorithms -- yielding the new research field of Deep Reinforcement Learning --  achieving outstanding performances in classic Atari games and in the Chinese board game Go \cite{mnih2015human,silver2017masteringgo}. In spite of some of these games having an intrinsic structured representation of the relations between their entities -- thus suggesting a symbolic approach -- their huge state space and reward flow limits the application of a sole DL model upon these representations. Therefore, the direct application of DL to symbolic domains is still in its lead-off stages \cite{garcez2015neural}.

The combination of connectionist and symbolic approaches may address the bottlenecks faced by these methods when they are used alone such as limited reasoning and hard-coding knowledge and rules \cite{bader2005,GarcezLG2009,KhardonR99}. In this sense, applying DL models to combinatorial problems arises as one of the main approaches towards achieving integrated machine learning (ML) and reasoning \cite{battaglia2018relational}. This family of problems do not show a simple mathematical structure but, in several cases, there are plenty of exact solvers available to them, which allows one to produce labelled datasets in any desired amount, even for DL models whose requirements for training data can be substantial. Moreover, they are naturally represented by graph structures. \cite{bengio2018machine} argue that such models, which combine combinatorial optimisation and ML techniques,  enhance the learning procedure mainly by: (1) dissecting the symbolic problem into smaller learning tasks and (2) exploring the space of decisions in search of the best performing behaviour. While (1) is due to the natural combinatorial optimisation structure, (2) is the main principle of all machine learning strategies.

One way of incorporating the relational structure of a combinatorial problem into a neural model is to ensure permutation invariance by letting adjacent elements of the problem communicate with themselves through neural modules subject to parameter sharing. That is, the problem \textit{prints} its graph representation onto the neural modules' configuration. These neural modules are primarily accountable for computing the messages sent among the problem's elements and for updating the internal representation of each of these elements. The family of models that makes use of this message-passing algorithm includes message-passing neural networks \cite{gilmer2017neural}, recurrent relational networks \cite{palm2017recurrent}, graph networks \cite{battaglia2018relational} and the pioneer model: graph neural network (GNN) \cite{gori2005new}.

Recently, \cite{selsam2018learning} developed a GNN solution to the $\mathcal{NP}$-Complete boolean satisfiability problem (SAT) which achieved around 85\% of accuracy on SAT instances containing 40 variables. On top of that, their model was able to decode satisfying assignments even though it was trained only to produce a boolean answer. \cite{selsam2018learning} also tackled other combinatorial problems reduced to SAT and were able to extract valid assignments in 85\% of the satisfiable instances. 

In this work, we introduce a model\footnote{Available at: \url{https://machine-reasoning-ufrgs.github.io/GNN-GCP/}} to tackle the decision version of the graph colouring problem (GCP), with no need of prior reductions. We also formalise our solution using a Graph Neural Network (GNN) framework \cite{gori2005new,scarselli2009graph,gilmer2017neural}. Our model mainly relies on GNN's power of coping with several types of edges as we approached the GCP problem by using two different types of vertices. By designing a GNN model to solve an important combinatorial problem (with applications on flow management \cite{Barnier2004}, job scheduling \cite{thevenin2018}, register allocation \cite{Chen2018} and others), we hope we can foster the adoption and further research on GNN-like models which in turn integrate deep learning and combinatorial optimisation. We believe that, from an AI perspective, our work provides useful insights on how neural modules reason over symbolic problems, and also on how their hidden states or embeddings can be interpreted.

The remainder of the paper is structured as follows. First, we  describe our solution to the Graph Colouring Problem in terms of a GNN-based model. We then report the model's performance on different datasets along with baseline heuristics comparisons. Finally, we present an analysis of results and directions for further research.

\section{A GNN Model for Decision GCP}
\label{GNNmodel}
Usually, graph neural network models assign multidimensional representations, or embeddings $\in \mathbb{R}^d$, to vertices and edges. These embeddings are then refined according to some adjacency information throughout a given number of message-passing iterations. The adjacency information controls which are the valid incoming messages for a given vertex (or edge), these filtered messages undergo an aggregating function and finally a Recurrent Neural Network (RNN) receives the aggregated messages and computes the embedding update for the given vertex. The Graph Network model \cite{battaglia2018relational} also allows the instantiation of global graph attributes  which seems appropriate to the $k$-colorability problem. However, as we treat each possible colour as an attribute, there must be multiple global graph attributes, i.e. each colour has its own embedding. Because of that, we chose to model the $k$-colorability problem in a Graph Neural Network framework \cite{gori2005new}, a seminal formalisation which already leveraged the capability of dealing with several types of nodes. GNN models were already proven to be promising on solving relational problems, even when dealing with numeric information, such as the travelling salesperson problem (TSP) \cite{prates2018}.

Given a GCP instance $I = (\mathcal{G},C)$ composed of a graph $\mathcal{G} = (\mathcal{V}, \mathcal{E})$ and a number of colours $C \in \mathbb{N}~|~C > 2$, each colour is assigned to a random initial embedding over an uniform distribution $\overset{(1)}{\mathbf{C}}[i] \sim \mathcal{U}(0,\,1) ~|~ \forall c_i  \in \mathcal{C}$ and the model initially assigns the same embedding $\in \mathbb{R}^d$ to all $V$ \emph{vertices}: this embedding is randomly initialised and then it becomes a trained parameter learned by the model. To allow the communication between neighbouring vertices and between vertices and colours, besides the vertex-to-vertex adjacency matrix $\mathbf{M_{\mathcal{VV}}} \in \{0,1\}^{|\mathcal{V}| \times |\mathcal{V}|}$, the model also requires a vertex-to-colour adjacency matrix $\mathbf{M_{\mathcal{VC}}} \in \{1\}^{|\mathcal{V}| \times |\mathcal{C}|}$, that connects each colour to all vertices since we chose to give no prior information to the model; i.e., a priori any vertex can be assigned to any colour. After this initialisation adjacent vertices and colours communicate and update their embeddings during a given number of iterations. Then the resulting vertex embeddings are fed into a MLP which computes a logit probability corresponding to the model's prediction of the answer to the decision problem: ``does the graph $G$ accept a $C$-coloration?''. This procedure is summarised in Algorithm \ref{alg:GNN-GCP}. 

\begin{algorithm}[t]
\begin{algorithmic}[1]
\Procedure{GNN-GCP}{$\mathcal{G} = (\mathcal{V},\mathcal{E}), C$}
\State
\LineComment{{\small Compute binary adjacency matrix from vertex to vertex}}
\State $\mathbf{M_{\mathcal{VV}}}[i,j] \negthickspace \leftarrow \negthickspace 1 \textrm{ iff } (\exists e \in \mathcal{E} | e \negthickspace = \negthickspace (v_i,v_j)) |~ \forall v_i \negmedspace \in \negmedspace \mathcal{V}, v_j \negmedspace \in \negmedspace \mathcal{V}$
\State
\LineComment{{\small Compute binary adjacency matrix from vertices to colours}}
\State $\mathbf{M_{\mathcal{VC}}}[i,j] \negthickspace \leftarrow \negthickspace 1 \forall v_i \negmedspace \in \negmedspace \mathcal{V}, c_j \negmedspace \in \negmedspace \mathcal{C}$
\State
\LineComment{{\small Compute initial vertex embeddings}}
\State $\overset{(1)}{\mathbf{V}}[i] \sim \mathcal{N}(0,\,1) ~|~ \forall v_i  \in \mathcal{V}$ 

\State
\LineComment{{\small Compute initial colour embeddings}}
\State $\overset{(1)}{\mathbf{C}}[i] \sim \mathcal{U}(0,\,1) ~|~ \forall c_i  \in \mathcal{C}$

\State
\LineComment{Run $t_{max}$ message-passing iterations}
\For{$t=1 \dots t_{max}$}
  \LineComment{{\small Refine each vertex embedding with messages received from its neighbours and candidate colours}}
  \label{alg:line:vertices_refinement}\State $\overset{(t+1)}{\mathbf{V}_h}, \overset{(t+1)}{\mathbf{V}} \negthickspace \leftarrow \negthickspace V_u(\overset{(t)}{\mathbf{V}_h}, \mathbf{M_{\mathcal{VV}}} \negthickspace \times \negthickspace \overset{(t)}{\mathbf{V}}, \negthickspace \mathbf{M_{\mathcal{VC}}} \times \negthickspace \underset{msg}{C}(\overset{(t)}{\mathbf{C}}))$
  \LineComment{{\small Refine each colour embedding with messages received from all vertices}}
  \label{alg:line:edges_refinement}\State $\overset{(t+1)}{\mathbf{C}_h}, \overset{(t+1)}{\mathbf{C}} \negthickspace \leftarrow \negthickspace C_u(\overset{(t)}{\mathbf{C}_h},\mathbf{M_{\mathcal{VC}}}^{T} \negthickspace \times \negthickspace \underset{msg}{V}(\overset{(t)}{\mathbf{V}}))$
\EndFor

\LineComment{{\small Translate vertex embeddings into logit probabilities}}
\State $V_{logits} \leftarrow V_{vote}\left(\overset{t_{max}}{\mathbf{V}}\right)$
\LineComment{{\small Average logits and translate to probability (the operator $\langle \rangle$ indicates arithmetic mean)}}
\State $\textrm{prediction} \leftarrow \textrm{sigmoid}(\langle \mathbf{V_{logits}} \rangle)$

\EndProcedure
\end{algorithmic}
\caption{Graph Neural Network Model for GCP}\label{alg:GNN-GCP}
\end{algorithm}

Our GNN-based algorithm updates vertex and colour embeddings, along with their respective hidden states, according to the following equations:

\begin{align}
\begin{split}\label{eq:1}
    \mathbf{V}^{(t+1)}, \mathbf{V}_h^{(t+1)} \leftarrow {}& \mathcal{V}_{u}(\mathbf{V}_h^{(t)}, \mathbf{M_{\mathcal{VV}}} \times (\mathbf{V}^{(t)}), \\
 &\hphantom{{}=\mathcal{V}_{u}} \mathbf{M_{\mathcal{VC}}} \times \underset{msg}{C}(\mathbf{C}^{(t)}))
\end{split}\\
\begin{split}\label{eq:2}
    \mathbf{C}^{(t+1)}, \mathbf{C}_h^{(t+1)} \leftarrow {}& \mathcal{C}_{u}(\mathbf{C}_h^{(t)}, \mathbf{M_{\mathcal{VC}}}^{\intercal} \times \underset{msg}{V}(\mathbf{V}^{(t)}))
\end{split}
\end{align}

In this case, the GNN model needs to learn the message functions (implemented via MLPs) $\underset{msg}{C}: \mathbb{R}^{d} \rightarrow \mathbb{R}^{d}$, which will translate colours embeddings into messages that are intelligible to a vertex update function, and $\underset{msg}{V}: \mathbb{R}^{d} \rightarrow \mathbb{R}^{d}$, responsible for translating vertices embeddings into messages. Also, it learns a function (RNN) responsible for updating vertices $V_{u}: \mathbb{R}^{2d} \rightarrow \mathbb{R}^{d}$ given its hidden state and received messages and another RNN to do the analogous procedure to the colors $V_{u}: \mathbb{R}^{2d} \rightarrow \mathbb{R}^{d}$.

\section{Training Methodology}
\label{train}

\begin{figure}[h]

    \centering
    \includegraphics[width=\linewidth]{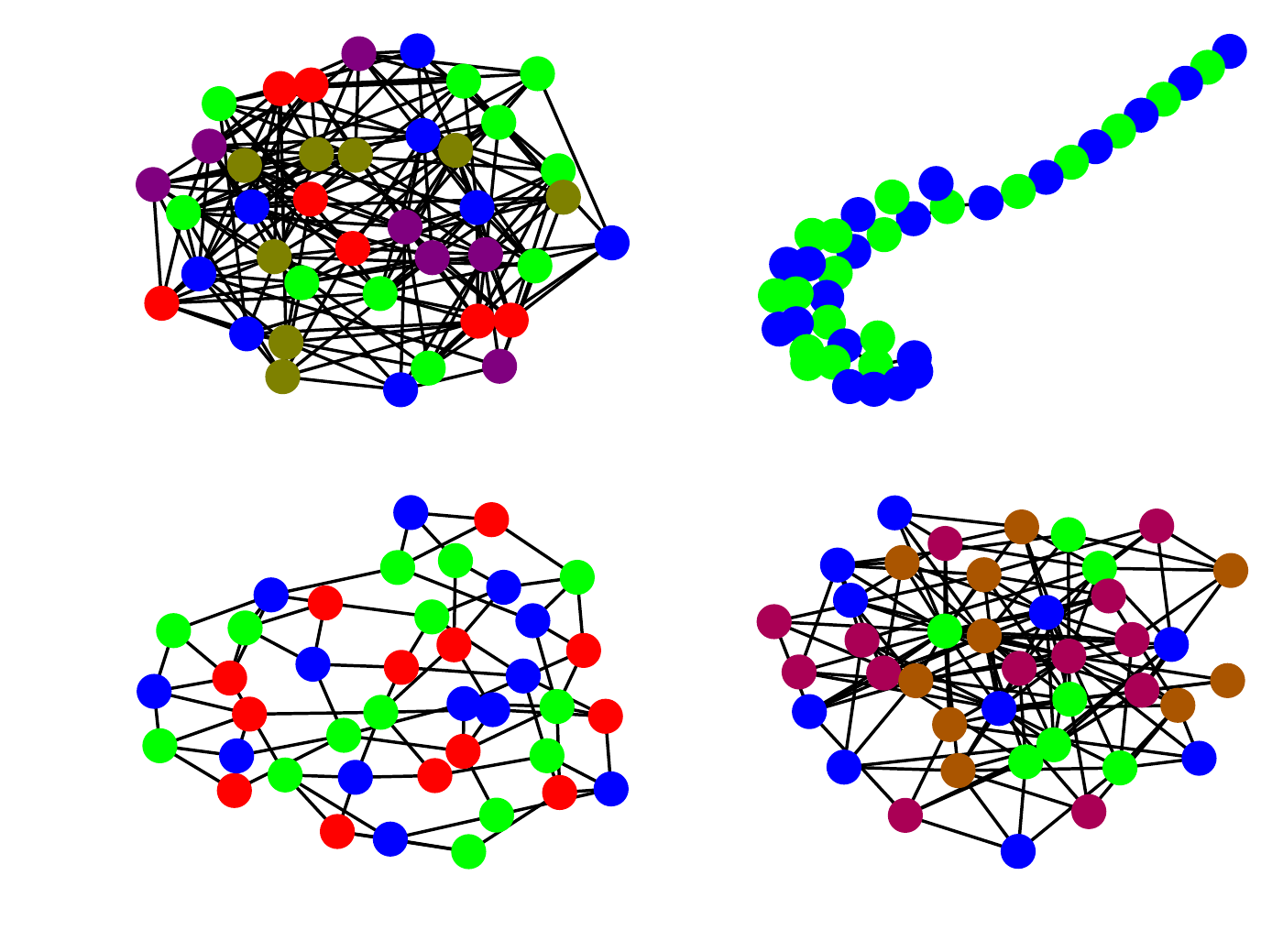}
    \caption{Pictorial representation of random training instances (top left) and some of the structured test instances (clockwise: power-law tree, power-law cluster, small-world). All instances are coloured with a number of colours equal to their chromatic number.}
    \label{fig:inst}
\end{figure}

To train these message computing and updating modules, MLPs and RNNs respectively, we used the Stochastic Gradient Descent algorithm implemented via TensorFlow's Adam optimiser. We defined as loss the binary cross entropy between the model final prediction and the ground-truth for a given GCP instance -- a boolean value indicating that the graph accepts (or not) the target colorability. We intended to train our model with very hard GCP instances. To do so, our training instances, with number of vertices $n \sim \mathcal{U}(40,60)$, were produced on the verge of phase transition: for each instance $I = (G=(\mathcal{V}, \mathcal{E}), C) | C = \chi(G)$\blfootnote{$\chi$ stands for chromatic number, i. e. the smallest value to obtain a k-colouring}, there is also an adversarial instance $\overline{I} = (G'=(\mathcal{V}, \mathcal{E'}), C) | C+1 = \chi(G')$ such that $E \neq E'$ only for a single edge $(v_i,v_j)$. This edge is usually called \emph{frozen edge} since for every valid colouring $C$ of $G$, $C[v_i] = C[v_j]$, which implies that this edge cannot belong to any $C$-colourable graph containing G as a subgraph \cite{culberson2001}. To produce a pair of instances, first we randomly chose a target chromatic number $\chi$ between 3 and 8, then populate the adjacency matrix $M_{VV}$ with a connectivity probability $p$ adjusted to the selected chromatic number, and finally a CSP-Solver\footnote{\url{https://developers.google.com/optimization/cp/cp_solver}} is used to ensure that the undirected graph represented by the initial matrix $M_{VV}$ has a chromatic number $\chi$: if it has, then we proceed adding edges to the graph until the CSP-Solver is no longer able to solve the GCP for $\chi$. The last two generated instances were added to the dataset (both considering $C = \chi$), ensuring that the dataset is not only composed by hard instances but also perfectly balanced: 50\% of the instances do not accept a $C$-colouring while the remaining 50\% accept. A total of $2\times2^{15}$ such instances were produced. An example of such training is depicted in Fig. \ref{fig:inst}.

These instances were randomly joined into a larger graph during training to produce a batch-graph containing $2 \times 8$ instances. This was done with a disjoint union so that the $M_{VV}$ and the $M_{VC}$ of each instance does not allow communication between vertices and colours of different subgraphs. Therefore, we ensure that despite being on the same batch-graph, there is no inter-graph communication. The logit probabilities computed for each vertex within the batch are separated and averaged according to which instance the vertex belongs to. Finally, we calculated the binary cross entropy between these predictions and their instances labels. Upon the initialisation, the model is defined with $64$-dimensional embeddings for vertices and colours. The MLPs responsible for message computing are three-layered (64,64,64) with ReLU nonlinearities as the activations for all layers except for the linear activation on the output layer. The RNN cells are basic LSTM cells with layer normalisation and ReLu activation. All results presented hereafter considered $T_{max} = 32$ time steps of message-passing.

\section{Experimental Results and Analyses}
\label{results}
\begin{figure*}[]
    \centering
    \includegraphics[width=\linewidth]{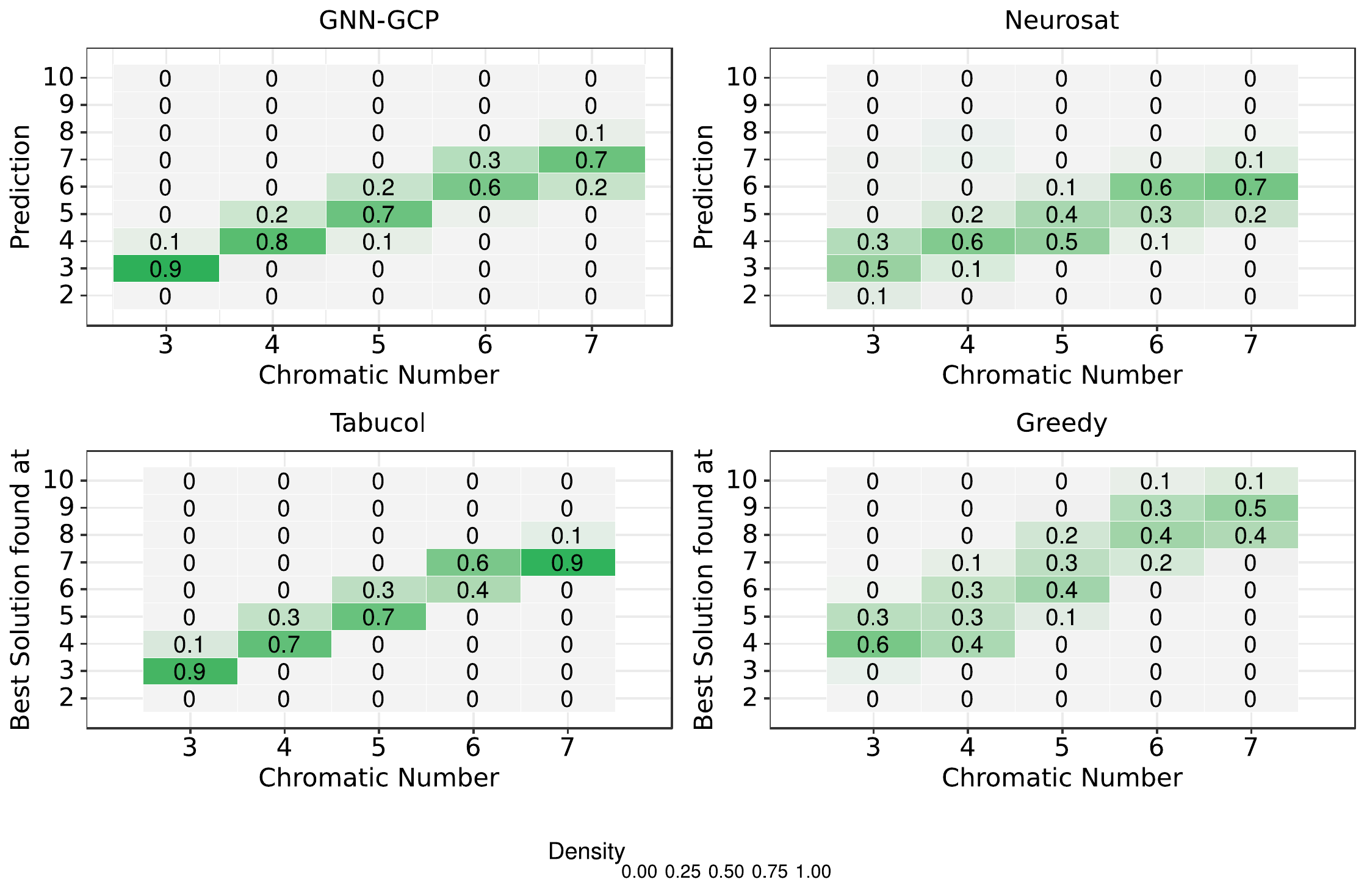}
    \caption{Prediction distributions over 4096 unseen test instances, with similar features to those seen in training, for our model (GNN), Neurosat, Tabucol and the Greedy algorithm. Note that the darker the main diagonal, the better the predictions. }
    \label{fig:testing-GNN-tabu}
\end{figure*}

We stopped the training procedure when the model achieved 82\% of accuracy and 0.35 of Binary Cross Entropy loss averaged over 128 batches containing 16 instances at the end of 5300 epochs. To produce the chromatic number on 4096 test instances (which followed the same distribution of the training ones) we fed the model with the same instance several times, with $C \in (2, \chi+3)$, and assigned as the chromatic number the first $C$ which implied in a positive answer from our model. Figure  \ref{fig:testing-GNN-tabu} shows how the GNN model performed over these test instances according to their chromatic number. Under the same settings, i.e. $d = 64$ and 32 timesteps, we trained our implementation of Neurosat with the same GNN-GCP training instances, but reduced to SAT instances. We also compared GNN-GCP's performance with two heuristics: Tabucol \cite{Hertz1987}, a local search algorithm which inserts single moves into a tabu list; and a greedy algorithm which assigns to a vertex the first available colour. As both heuristics outcomes are valid colouring assignments, they never underestimate the chromatic number, as opposed to our model.

Along the first six chromatic numbers our model slightly outperforms the Tabucol algorithm w.r.t. accuracy -- hit only when the exact chromatic number is achieved -- however, it demonstrates a drop on its performance at the highest chromatic number, where the Tabucol achieves around 90\% of accuracy.  Nevertheless, our model's absolute deviation from the exact chromatic number averaged 0.2473, while the Tabucol and the greedy algorithm achieved 0.2878 and 1.5544, respectively.

\begin{figure}[h]
    \centering
    \includegraphics[width=\linewidth]{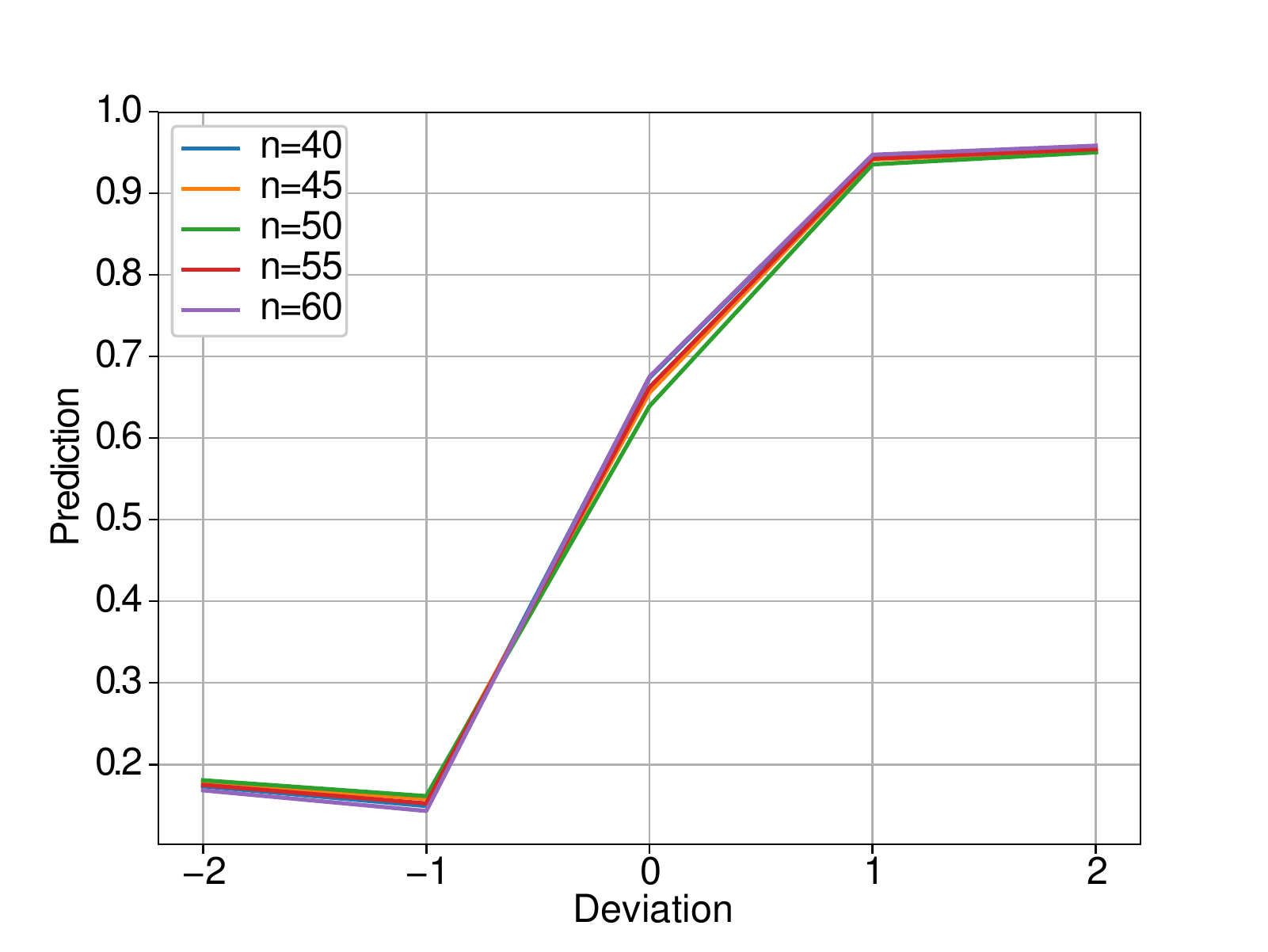}
    \caption{ Average prediction -- above 0.5 means a positive answer -- extracted from the model for testing instances with size ranging from 40 to 60. Each of the 128 instances were fed seven times to the model with target $\mathcal{C} \in [\chi - 2, \chi + 2]$  }
    \label{fig:accept-curve}
\end{figure}

Even though we trained our model on instances whose $C$ was true ($C = \chi$) or false ($C = \chi-1$) only by a narrow margin, we could only state it can solve the GCP problem if as the margin goes wider, the model still gives the right answer, despite being positive or negative. We can see that in Fig.  \ref{fig:accept-curve}: our model's predictions undergo a regime remindful of a phase transition -- as we fed our model with $C$ values closer to $\chi$ it became unsure about its prediction, nevertheless the model is quite sure about its prediction on the upper ($C = \chi+2$) and lower ($C = \chi-2$) bounds. Phase transition is a well-known phenomenon in several combinatorial problems, including SAT \cite{dudek2017combining} and the GCP itself \cite{lenka2007}, in which the existence and the distribution of colouring solutions are related to several levels of connectivity within an instance. We exploited this feature to generate our training instances and our model also presents a similar behaviour during the test.

\subsection{Performance on Real and Structured Instances}

In order to assess the performance of our trained model in unseen instances, both larger and smaller than those it was trained upon, we gathered 20 instances from the COLOR02/03/04 Workshop dataset\footnote{\url{https://mat.tepper.cmu.edu/COLOR02/}} and fed them to our model and to the previously cited heuristics (Tabucol and Greedy). These instances have sizes (number of vertices) up to 835\% larger than the training ones, and also have chromatic numbers exceeding the boundaries seen during the training procedure. We also fed them to our Neurosat model, but it was not able to output a positive answer inside the range of $[2,\chi+5]$ for all instances. This suggests that these different graph distributions are not well suited for the trained Neurosat model. It is also worth mentioning that to reduce from GCP to SAT one is required to create $C*\mathcal{V}$ variables, $C*\mathcal{V}$ clauses to ensure that there will be no uncoloured vertex, $C*\mathcal{E}$ clauses to ensure that each edge has its source and target coloured differently and $(C-1)*\mathcal{V}$ clauses to ensure that at most one colour will be assigned to each vertex, thus causing a significant increase of nodes and edges in the resulting SAT graph in comparison to the original GCP graph.

\ifx
\begin{figure*}[b]
    \centering
    \includegraphics[width=\linewidth]{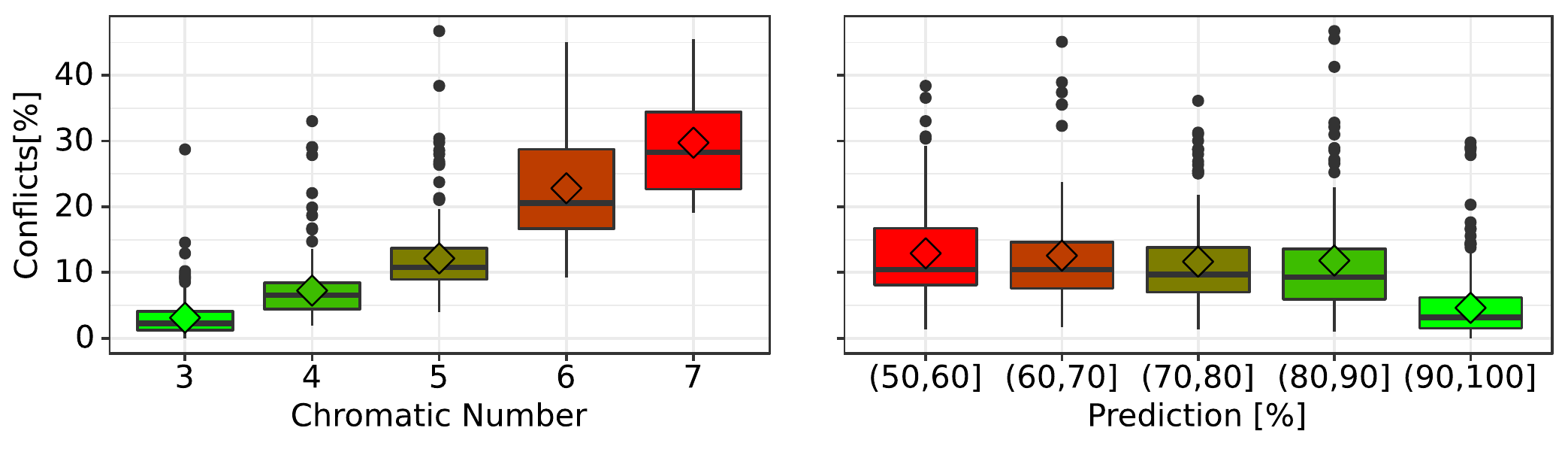}
    \caption{After performing a k-means algorithm on the vertex embeddings, we computed the ratio of conflicts (adjacent vertices pertaining to the same cluster) within each cluster. In this experiment, we fed our model with the exact chromatic number for each instance and selected the embeddings only for predictions above 50\%: GNN is positive that there is a colouration. In the left graph one can see that the clusters have less meaning as the chromatic number grows. The rightmost graph shows how the clustering correlates to the GNN final prediction -- when the model is more confident that there is a colouration, the clusters have less conflicts. }
    \label{fig:clustering}
\end{figure*}
\fi

To compute the chromatic number we adopted the same previous procedure: with $C$ ranging from $C = 2$ up to $C = \chi + 3$, the first $C$ which yielded a positive prediction/answer is considered the chromatic number. If our model -- or the Tabucol -- did not present a positive answer until this upper boundary, it accounts for a NA answer on Table \ref{tab:real-instances}.

As both Tabucol and the greedy algorithm produce a valid colouring assignment, they never underestimate the chromatic number, as our model eventually does. When it comes to predict the exact chromatic number, our model only achieved 4 hits, against 11 and 7 from the Tabucol and the Greedy algorithm, respectively. Nevertheless, our model's absolute deviation from the actual chromatic number accounted for 1.15, standing in between Tabucol (0.33) and greedy algorithm (2.15). Despite its overall low performance on these instances, it is worthy reminding that upon training our model have never seen instances with more than 60 vertices and $\chi$ larger than 7.

We also tested our model on three different random graphs distributions, namely: random power-law tree ($\gamma = 3$), Watts-Strogatz small-world \cite{watts1998} ($k = 4, p = 0.25$) and Holme and Kim model \cite{holme2002} ($m = 4, p = 0.1$). We generated 100 instances of each distribution with size ranging from 32 to 128 vertices. Examples of such instances are depicted in Fig. \ref{fig:inst} (top right, bottom left and bottom right). By their own definition, all power-law tree graphs are 2-coloured, a target $C$ never seen by our model during training, and have their vertices ordered by degree, thus allowing the greedy algorithm to achieve a perfect performance. The same happens on the Holme-Kim distribution whose graphs are built through preferential attachment and also has a power-law distribution, thus there is a high probability that a selected vertex has a low degree as well as a high probability that it is connected to a vertex of high degree, there is also the fact that the vertices have a probability of creating a triangle, thus constraining the colours that can be picked after selecting a vertex. With this in mind, the algorithm can easily assign colours greedily throughout the low degree vertices and assign differing colours to high degree ones when they are reached, which also constrains the bridges between these vertices.
\begin{table}[]
\small
\centering
\caption{The chromatic number produced by our model and two heuristics on some instances of the COLOR02/03/04 dataset. As our model faces unseen graph sizes and larger chromatic numbers it tends to underestimate its answers.}
\label{tab:real-instances}
\begin{tabular}{cccccc}
\hline
\multirow{2}{*}{Instance} & \multirow{2}{*}{Size} & \multirow{2}{*}{$\chi$} & \multicolumn{3}{c}{Computed $\chi$} \\
 &  &  & \multicolumn{1}{c|}{GNN} & \multicolumn{1}{c|}{Tabucol} & Greedy \\ \hline
queen5\_5 & 25 & 5 & 6 & \textbf{5} & 8 \\
queen6\_6 & 36 & 7 & \textbf{7} & 8 & 11 \\
myciel5 & 47 & 6 & 5 & \textbf{6} & \textbf{6} \\
queen7\_7 & 49 & 7 & 8 & 8 & 10 \\
queen8\_8 & 64 & 9 & 8 & 10 & 13 \\
1-Insertions\_4 & 67 & 4 & \textbf{4} & 5 & 5 \\
huck & 74 & 11 & 8 & \textbf{11} & \textbf{11} \\
jean & 80 & 10 & 7 & \textbf{10} & \textbf{10} \\
queen9\_9 & 81 & 10 & 9 & 11 & 16 \\
david & 87 & 11 & 9 & \textbf{11} & 12 \\
mug88\_1 & 88 & 4 & 3 & \textbf{4} & \textbf{4} \\
myciel6 & 95 & 7 & \textbf{7} & \textbf{7} & \textbf{7} \\
queen8\_12 & 96 & 12 & 10 & \textbf{12} & 15 \\
games120 & 120 & 9 & 6 & \textbf{9} & \textbf{9} \\
queen11\_11 & 121 & 11 & 12 & NA & 17 \\
anna & 138 & 11 & \textbf{11} & \textbf{11} & 12 \\
2-Insertions\_4 & 149 & 4 & \textbf{4} & 5 & 5 \\
queen13\_13 & 169 & 13 & 14 & NA & 21 \\
myciel7 & 191 & 8 & NA & \textbf{8} & \textbf{8} \\
homer & 561 & 13 & 14 & \textbf{13} & 15 \\ \hline
\end{tabular}
\end{table}

\begin{table}[]
\scriptsize
\caption{Strict accuracy of our model and the two algorithms considering three random graphs distributions}
\label{tab:diff-distrib}
\resizebox{\linewidth}{!}{%
\begin{tabular}{@{}cclclcl@{}}
\toprule
\multirow{2}{*}{Distribution} & \multicolumn{2}{c}{GNN} & \multicolumn{2}{c}{Tabucol} & \multicolumn{2}{c}{Greedy} \\
 & \multicolumn{2}{c|}{Accuracy {[}\%{]}} & \multicolumn{2}{c|}{Accuracy {[}\%{]}} & \multicolumn{2}{c}{Accuracy {[}\%{]}} \\ \midrule
Power Law Tree & \multicolumn{2}{c}{100.0} & \multicolumn{2}{c}{93.0} & \multicolumn{2}{c}{100.0} \\
Small-world & \multicolumn{2}{c}{90.0} & \multicolumn{2}{c}{77.0} & \multicolumn{2}{c}{9.0} \\
Holme and Kim & \multicolumn{2}{c}{54.1} & \multicolumn{2}{c}{76.4} & \multicolumn{2}{c}{100.0} \\ \bottomrule
\end{tabular}%
}
\end{table}

\subsection{Exploring vertex embeddings}
The capability of GNN-like models to generate meaningful embeddings and to learn an algorithm to solve its task was already highlighted by \cite{selsam2018learning}.
Following their findings, even though we trained our model only to produce a boolean answer, we also expected to decode valid colouring assignments to each vertex. To achieve that, we extracted vertex embeddings from 1024 test instances which were fed to our model along with their exact chromatic number as $C$ and resulted in a GNN prediction $> 50\%$. These embeddings were then clustered into $C$ groups -- we made the \textit{a priori} assumption that the model internally places non-adjacent vertices near each other, thus resulting into colour assignments. For each cluster of each instance we computed how many conflicts were raised -- the ratio between the amount of adjacent vertices pertaining to that cluster and the number of 2-combinations without repetition of these vertices. We then computed the average conflicts per cluster/colour. Naturally, a perfect valid colouring assignment for a given instance would imply in zero conflicts.
\ifx
\begin{figure*}[t]
    \centering
    \includegraphics[width=\linewidth]{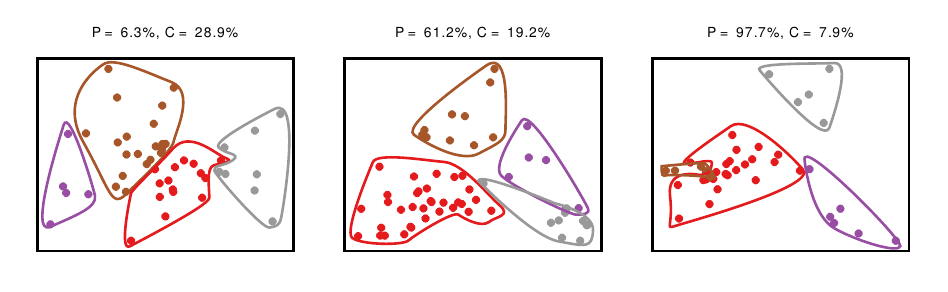}
    \caption{Vertex embeddings (after a PCA-2D procedure) of three different test instances, with $\chi=4$. The axes and the surrounding curves have no meaning as we are simply interested in visualising how the clusters behaviour are related to our model outcomes. All these three instances should imply in a positive answer, but our model only answered positively to the second and to the third one. }
    \label{fig:clustering-ex}
\end{figure*}
\fi

\begin{figure*}[t]
    \centering 
    \includegraphics[width=\linewidth]{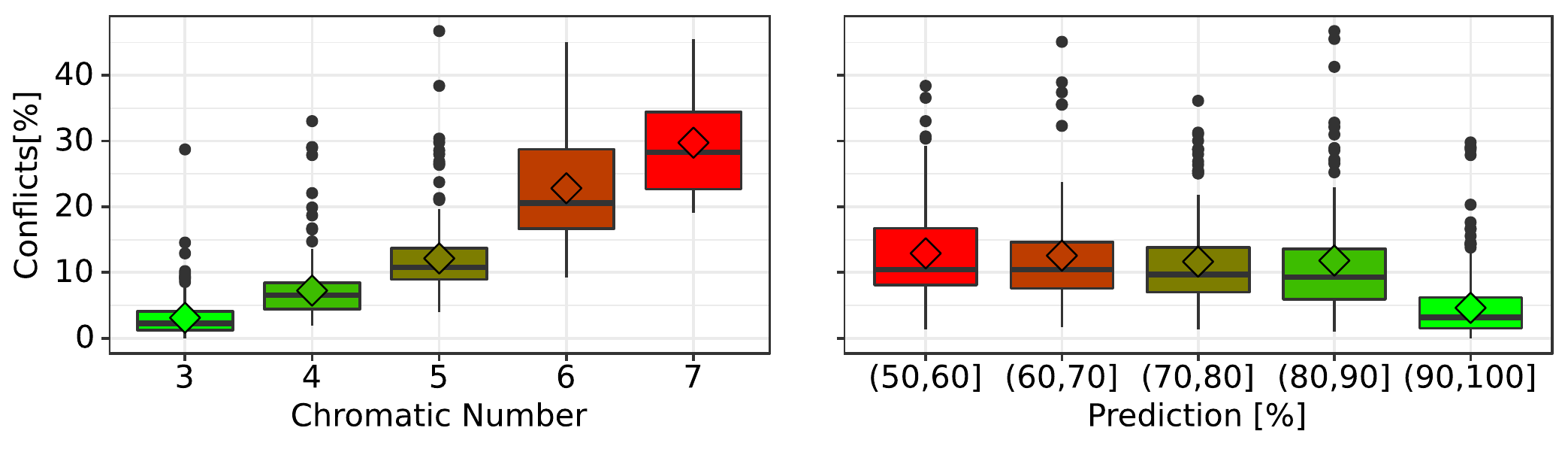}
    \caption{After performing a k-means algorithm on the vertex embeddings, we computed the ratio of conflicts (adjacent vertices pertaining to the same cluster) within each cluster. In this experiment, we fed our model with the exact chromatic number for each instance and selected the embeddings only for predictions above 50\%: GNN is positive that there is a colouration. In the left graph one can see that the clusters have less meaning as the chromatic number grows. The rightmost graph shows how the clustering correlates to the GNN final prediction -- when the model is more confident that there is a colouration, the clusters have less conflicts. }
    \label{fig:clustering}
\end{figure*}
According to Fig. \ref{fig:clustering} (leftmost boxplots) our model faces more difficulties in assigning correct vertices to clusters when $C$ grows -- regarding instances with $C = 7$ the average conflicts ratio within the seven clusters achieved around 30\%.

On the other hand, this metric drops off to below 5\% regarding 3-colouring instances. We also verified a moderate negative correlation (-0.6 -- Spearman's Rank Correlation) between the positive certainty of our model and the amount of conflicts, as seen in Fig. \ref{fig:clustering} (rightmost boxplots), which goes along with the natural thought that the fewer the conflicts within each instance's clusters, the more certain our model is of a positive answer. Among the 1024 instances, our model together with the clustering procedure were able to provide a valid colouring assignment to only 3 examples, nevertheless its behaviour suggests that it is able to respond positively when some inferior threshold w.r.t. conflicts is reached.

Figure \ref{fig:clustering-ex} helps visualising how the vertices' distribution along the dimensions (internal outcome of our model) and how their clustering affects the GNN predictions. The clusters separability (measured with a silhouette score $S \in [-1,+1]$ over the 64-dimensional clustered embeddings) increases together with our model's certainty of a positive answer and the inverse ratio of conflicts within each cluster: the leftmost example resulted in a clustering with 28.9\% of average conflicts and a $S$ equals to 0.29, even though our model should have answered it positively, its prediction was only 6.3\% (any answer below 50\% is considered negative); when fed with the middle instance, however, our model was able to answer properly (61.2\%) as not only the conflicts' average decreased to 19.2\% but also the silhouette coefficient increased to 0.39; finally, in the last example our model was quite sure about its positive answer, which goes along with the clustering procedure outcomes: only 7.9\% of conflicts within each cluster and a silhouette score of 0.44.
\begin{figure*}[t]
    \centering
    \includegraphics[width=\linewidth]{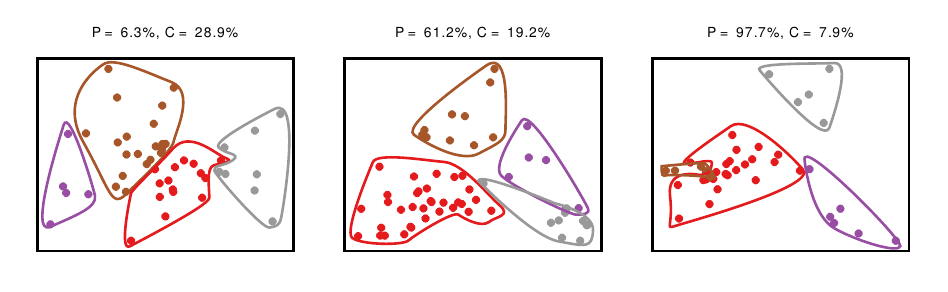}
    \caption{Vertex embeddings (after a PCA-2D procedure) of three different test instances, with $\chi=4$. The axes and the surrounding curves have no meaning as we are simply interested in visualising how the clusters behaviour are related to our model outcomes. All these three instances should imply in a positive answer, but our model only answered positively to the second and to the third one. }
    \label{fig:clustering-ex}
\end{figure*}

\section{Conclusions and Future Work}
\label{conc}

In this paper, we have shown how GNN models effectively tackle the Graph Colouring Problem. Our model works by keeping and updating a memory containing high dimensional representations of both vertices and colours. After 32 message-passing iterations between adjacent vertices and between vertices and colours each vertex voted for a final answer on whether the given graph admits a $C$-colouring. We trained the model following an adversarial procedure: for each positive instance, we also produced a very similar one which differs only by an addition of a single edge, causing an increase of the original $\chi$ chromatic number; both these instances were fed to our model with $C$ equals to the $\chi$ value of the positive instance, yielding a perfectly balanced and hard dataset. Upon training our model achieved 82\% accuracy with instances with $N \in [40,60]$ and $C \in [3,7]$. We also demonstrated how this trained model was able to generalise  its results to previously unseen target $C$ and structured and larger instances, yielding a performance comparable to a well-know heuristic (Tabucol). In spite of being trained on the verge of satisfiability, we also showed a curve depicting how our model behaved to varying values of the target colour $C$ higher or lower than its chromatic number (Figure \ref{fig:accept-curve}). 

Finally, we postprocessed the vertex embeddings to understand how they were inducing the final answer. We believe that internally our model searches for a positive answer by attempting to cluster vertices so that vertices that could have the same colour are near to each other. We uncovered some kind of threshold search as our model answers positively even when these clusters contain adjacent vertices, but in a low ratio.
We chose to present the target number of colours $C$ to the model by feeding it with random initialised $C$ embeddings so no prior knowledge is given, but these initial embeddings could be arranged in some other manner. Another future improvement is to attempt to minimise conflicts within the model itself (as a loss measure). This, we believe, could lead to better and more meaningful results despite the time and space complexity being increased. Nevertheless, we expect that our work can demonstrate how a GNN-like model can be manipulated into solving hard combinatorial problems such as GCP in an interpretable fashion, whose answers are not only accurate, but also constructive.

\section*{Acknowledgments}

We would like to thank NVIDIA for the GPU granted to our research group and also to Moshe Vardi for several suggestions and conversations that contributed to this research.


\bibliographystyle{IEEEtran}  
\bibliography{bibl} 

\end{document}